\documentclass[10pt,twocolumn,letterpaper]{article}
\usepackage[T1]{fontenc}
\usepackage[]{cvpr}

\def\confName{CVPR}
\def\confYear{2027}

\usepackage{amsmath,amsfonts,bm}

\def\eqref#1{equation~\ref{#1}}

\def\1{\bm{1}}

\def\vzero{{\bm{0}}}

\def\vc{{\bm{c}}}

\def\vm{{\bm{m}}}
\def\vn{{\bm{n}}}

\def\vu{{\bm{u}}}
\def\vv{{\bm{v}}}

\def\vx{{\bm{x}}}
\def\vy{{\bm{y}}}
\def\vz{{\bm{z}}}

\def\mA{{\bm{A}}}

\def\mG{{\bm{G}}}
\def\mH{{\bm{H}}}
\def\mI{{\bm{I}}}

\DeclareMathAlphabet{\mathsfit}{\encodingdefault}{\sfdefault}{m}{sl}
\SetMathAlphabet{\mathsfit}{bold}{\encodingdefault}{\sfdefault}{bx}{n}

\usepackage{amsmath,amssymb,amsthm,mathtools}
\usepackage{booktabs,multirow}
\usepackage{array,tabularx}
\usepackage{placeins}
\usepackage{stfloats}
\usepackage[nospread,noshrink]{flushend}
\usepackage{graphicx}
\DeclareUnicodeCharacter{3001}{\includegraphics{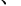}}
\DeclareUnicodeCharacter{203A}{\includegraphics{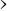}}

\usepackage{xcolor}
\usepackage{algorithm}
\usepackage{algorithmic}
\usepackage{tikz}
\usetikzlibrary{arrows.meta,positioning,calc}
\usepackage{pgfplots}
\pgfplotsset{compat=1.17}
\usepackage{microtype}
\definecolor{cvprblue}{rgb}{0.21,0.49,0.74}
\usepackage{xr-hyper}
\usepackage[pagebackref,breaklinks,colorlinks,allcolors=cvprblue]{hyperref}

\newcolumntype{L}{>{\raggedright\arraybackslash}X}
\makeatletter
\newcommand{\dorsneedspace}[1]{%
  \par\begingroup
  \@tempdima=\pagegoal
  \advance\@tempdima by -\pagetotal
  \ifdim\@tempdima<#1\relax\vfil\break\fi
  \endgroup
}
\makeatother
\newcommand{\xh}{\hat{\mathbf{x}}}

\newtheorem{proposition}{Proposition}

\theoremstyle{remark}

\newcommand{\dci}[2]{\textcolor{black!50}{\scriptsize[#1,\,#2]}}

\definecolor{tablebest}{HTML}{FF0000}
\definecolor{tablesecond}{HTML}{00B0F0}
\definecolor{tablehead}{HTML}{E6E6E6}
\definecolor{tableours}{HTML}{FFF8F1}
\newcommand{\best}[1]{\textcolor{tablebest}{\textbf{#1}}}
\newcommand{\second}[1]{\textcolor{tablesecond}{\textbf{#1}}}
\providecommand{\veps}{{\bm{\epsilon}}}

\newcommand{\ldiff}{\ell_{\mathrm{diff}}}
\newcommand{\lphys}{\ell_{\mathrm{phys}}}

\newcommand{\ourrow}{\textbf{DORS} (ours)}
\title{DORS: A Compact Dual-Observation Reverse Solver for Linear Inverse Problems}

\author{Fengyang Xiao$^{1,*}$\,,
Jingjia Feng$^{1,*}$\,,
Xinge Guo$^{1}$\,,
Zheng Chen$^{2,\dagger}$\,,
Baoxi Feng$^{1}$\,, \\
Guanyi Qin$^{3}$\,, 
            Chunming He$^{1,\dagger}$\,, 
        and {Sina Farsiu}$^{1}$\\
        $^1$Duke University,
	$^2$Shanghai Jiao Tong University,  $^3$National University of Singapore
\\
 $\dagger$ Corresponding Author, \quad
Contact: fengyang.xiao@duke.edu
}

\raggedend

\AtBeginDocument{%
  \setlength{\abovedisplayskip}{8pt plus 2pt minus 3pt}%
  \setlength{\belowdisplayskip}{8pt plus 2pt minus 3pt}%
  \setlength{\abovedisplayshortskip}{0pt plus 2pt}%
  \setlength{\belowdisplayshortskip}{5pt plus 2pt minus 3pt}%
}

\makeatletter
\AtEndDocument{%
  \@for\dorscounter:=equation,figure,table,proposition,remark,lemma,corollary,algorithm,linenumber\do{%
    \@ifundefined{c@\dorscounter}{}{% linenumber exists only in review mode
    \immediate\write\@auxout{\string\newlabel{dors:counter:\dorscounter}{{\number\value{\dorscounter}}{}{}{}{}}}}%
  }%
}
\makeatother
\usepackage{pifont}
\usepackage{colortbl}
\PassOptionsToPackage{table}{xcolor}

\begin{document}

\maketitle

\begin{abstract}
Conditional diffusion methods for linear inverse problems learn a reverse estimator predicting the clean image from a diffusion latent and a physical measurement.  
We present DORS, a compact dual-observation deterministic solver for linear inverse problems whose learned refiner can be reused across
measurement operators.
Derived from a residual diffusion path, DORS alternates analytic dual-observation fusion with a shared 3.7M-parameter refiner trained from scratch through the complete reconstruction chain.
The forward process specifies the latent’s observation model, so at every step the clean image is observed twice, through the current latent and original measurement, and both likelihoods are Gaussian with known parameters.
The refiner receives only the fused estimate and a scalar proximal coefficient, while the measurement model enters through initialization and fusion.
We show the latent quadratic is centered at a schedule-weighted average of earlier reconstructions, the trajectory center, whose influence the measurement operator attenuates in well-observed directions.
With matched backbone and training budgets, DORS improves over latent-conditioned refinement on motion deblurring and single-coil MRI.
Task-specific models outperform the compared methods on deblurring, inpainting, super-resolution, and MRI, with the fewest parameters and lowest measured peak memory in the deblurring comparison.
With refiner weights fixed, a deblurring model transfers to super-resolution and a single-coil MRI model to multi-coil reconstruction, the latter with one calibrated fusion scalar.
Code will be released.
\end{abstract}

\section{Introduction}
\label{sec:intro}

Linear inverse problems ask for a clean image $\vx_0\in\mathbb{R}^n$ given a physical measurement
\begin{equation}\label{eq:measurement}
    \vy=\mH\vx_0+\vn,\quad \vn\sim\mathcal{N}(\vzero,\sigma_y^2\mI),
\end{equation}
where $\mH$ is a known linear operator. Diffusion models address this problem along two routes: zero-shot methods keep an unconditional prior fixed and inject measurement information at inference through guidance or projections \citep{kawar2022ddrm,chung2023dps,song2023pigdm,wang2023ddnm,zhu2023diffpir,lugmayr2022repaint}. Supervised methods instead train on paired data, either as condition-injected denoisers \citep{saharia2023image,saharia2022palette,whang2022deblurring} or as stochastic paths that connect the degraded and clean domains directly \citep{luo2023irsde,delbracio2023indi,liu2023i2sb,yue2023resshift,zhou2024ddbm}. Supervised paths deliver strong few-step reconstructions, but as Sobieski et al.~\citep{sobieski2025sdb} observe, they have historically discarded the measurement model's structure. Recent work makes the measurement model explicit in supervised reconstruction \citep{sobieski2025sdb}, including through unrolled updates \citep{wang2025diffunfolding} and operator-conditioned feature processing \citep{elata2025invfussion}.

At each reverse step, the estimator receives the diffusion latent $\vx_t$ and the measurement $\vy$.
Let $\mG$ be a fixed lifting operator from the measurement to the image domain. Under the residual forward model, the shifted latent satisfies
\begin{equation}\label{eq:zt-intro}
    \vz_t
    = \vx_t-(1-\gamma_t)\mG\vy
    = \gamma_t\vx_0+\tau_t\veps,
    \quad
    \veps\sim\mathcal{N}(\vzero,\mI).
\end{equation}
Both \cref{eq:measurement} and \cref{eq:zt-intro} relate the same image $\bm{x}_0$ to an observation through a known linear operator and Gaussian noise, which specifies the two quadratic observation terms in our fusion.
 The diffusion schedule specifies $\gamma_t$ and $\tau_t$, while the measurement model specifies $\bm{H}$ and $\sigma_y$. 
 UD$^{2}$M~\citep{mbakam2025ud2m} and EPS~\citep{mammadov2026eps} combine the measurement
and diffusion likelihoods in a joint proximal update.
We study this quadratic structure in a compact deterministic
solver, where no noise is drawn, and the latent becomes a
state propagated from earlier reconstructions.

Should this state enter the learned module as a conditioning input or the analytic update? And in the analytic update, does the benefit come from the added quadratic curvature or from where the quadratic is centered?

We present DORS, a compact deterministic dual-observation reverse solver (Fig.~\ref{fig:framework}).
Each reverse step alternates \emph{analytic precision fusion}, one quadratic solve over the two likelihood terms and a proximal term, with \emph{shared learned refinement} by a 3.7M-parameter network that takes the fused estimate and a scalar proximal coefficient. Its weights are shared across all reverse steps and inner stages and trained from scratch through the same deterministic chain used at inference with reconstruction supervision at each reverse step and no noise injected into the reverse chain.

On this chain, the latent term is a quadratic centered at a schedule-weighted average of earlier reconstructions. This \emph{trajectory center} is propagated between reverse steps and held fixed during each step's inner stages. We characterize its effect at fixed quadratic curvature: the measurement operator filters the correction toward the center, attenuating it more strongly in directions with stronger measurement constraints. Analytic fusion suits this regime: the center is an image with a known weight, so the quadratic solve uses it exactly, together with $\mH$, $\vy$ and the timestep, and none of them enters the refiner. 
The measurement model remains explicit in the initialization and fusion, preserving the refiner's input interface when $\mH$ changes. With fixed refiner weights, we transfer a deblurring-trained model to super-resolution and a single-coil MRI model to multi-coil reconstruction, and the latter uses one calibrated scalar in the fusion (Sec.~\ref{sec:transfer_dors}). 

In controlled studies, analytic state handling improves PSNR over latent-conditioned refinement on motion deblurring and single-coil MRI at $4\times$. It keeps the curvature, but centering the term at the current iterate falls behind DORS. In the reported deblurring comparison, DORS also has the fewest parameters and the lowest measured peak memory.

Our contributions are summarized as follows.
\begin{itemize}[leftmargin=1.5em]
    \item \textbf{Compact deterministic solver.} We develop DORS, which alternates analytic dual-observation fusion with
    a shared 3.7M-parameter refiner trained from scratch
    through the full reconstruction chain.
    The measurement model enters initialization and fusion;
    the refiner takes only the fused estimate and a scalar
    proximal coefficient.
\item \textbf{Trajectory-center mechanism.}
DORS's deterministic recursion centers the latent
quadratic at a schedule-weighted average of earlier
reconstructions, fixed within each reverse step.
At fixed curvature, we derive how the measurement operator
filters corrections toward this center.
Controls with matched backbone and training budgets support
analytic state handling over learned conditioning and
center placement beyond added curvature.
\item \textbf{Evaluation and transfer.}
    Task-specific models outperform
    compared methods on four inverse problems, with the
    fewest parameters and lowest measured peak memory in
    the deblurring comparison.
    We show transferability across operators and tasks
    with fixed refiner weights.

\end{itemize}

 \section{Related Work}
 \label{sec:related}

 \noindent\textbf{Diffusion models for inverse problems.}
Diffusion models support both zero-shot and supervised
image restoration~\cite{he2025diffusionsurvey}.
Zero-shot solvers combine a pretrained image prior with
the measurement model through posterior
guidance~\cite{chung2023dps,song2023pigdm},
spectral updates~\cite{kawar2022ddrm},
measurement-consistency projections~\cite{wang2023ddnm},
splitting with diffusion denoisers~\cite{zhu2023diffpir},
or decoupled noise annealing~\cite{zhang2025daps}.
Supervised methods learn reverse estimators from paired
data, using measurement-conditioned
denoisers~\cite{saharia2022palette,saharia2023image,whang2022deblurring}
or residual and bridge paths connecting degraded and
clean images~\cite{delbracio2023indi,liu2023i2sb,luo2023irsde,yue2023resshift,zhou2024ddbm}.
Reti-Diff~\cite{he2025retidiff} generates compact
reflectance and illumination priors through latent
diffusion to guide restoration.
DORS adopts a residual diffusion path and trains a
deterministic reverse chain with reconstruction supervision.

\noindent\textbf{Model-based reconstruction with learned priors.}
Deep unfolding turns optimization iterations into
trainable network stages~\cite{gregor2010lista}.
Task-specific formulations extend this approach to
degraded image fusion~\cite{he2023derun},
real-world dehazing~\cite{fang2024corun},
and illumination restoration~\cite{he2026unfoldir}.
DSDNet+~\cite{kuo2025efficient} solves the data step in
closed form, while MoDL~\cite{aggarwal2019modl} shares
one network across iterations and enforces data consistency
with conjugate gradients.
Plug-and-play methods replace the prior update with
a denoiser~\cite{venkatakrishnan2013pnp},
while DPIR~\cite{zhang2021dpir} reuses a
noise-level-conditioned denoiser across restoration problems.
Memory modules and feature connections also pass
information across unfolding
stages~\cite{song2021madun,mou2022dgunet}.
DORS keeps the quadratic data update and learned image
update separate, adding the latent term to the data update.
Along the reverse chain, this term carries earlier
reconstructions into later fusions, with aggregation
determined by the schedule and reverse recursion.

\noindent\textbf{Operator-aware reverse estimators.}
CDDB~\cite{chung2023cddb} adds data-consistency corrections
to direct diffusion bridges, while
Diff-Unfolding~\cite{wang2025diffunfolding} and
InvFusion~\cite{elata2025invfussion} incorporate the operator
through unrolling and operator-aware feature processing,
respectively.
UD$^{2}$M~\citep{mbakam2025ud2m} jointly handles the
measurement and diffusion likelihoods through a proximal
update within a posterior sampler adapted from a pretrained
diffusion model.
DORS studies the same quadratic fusion structure in a
compact deterministic reconstruction chain, with a refiner
trained from scratch and a propagated state that admits
a trajectory-center representation.

\section{DORS: Dual-Observation Reverse Solver}
\label{sec:method}

\begin{figure*}[!t]
\centering
\includegraphics[width=1\textwidth]{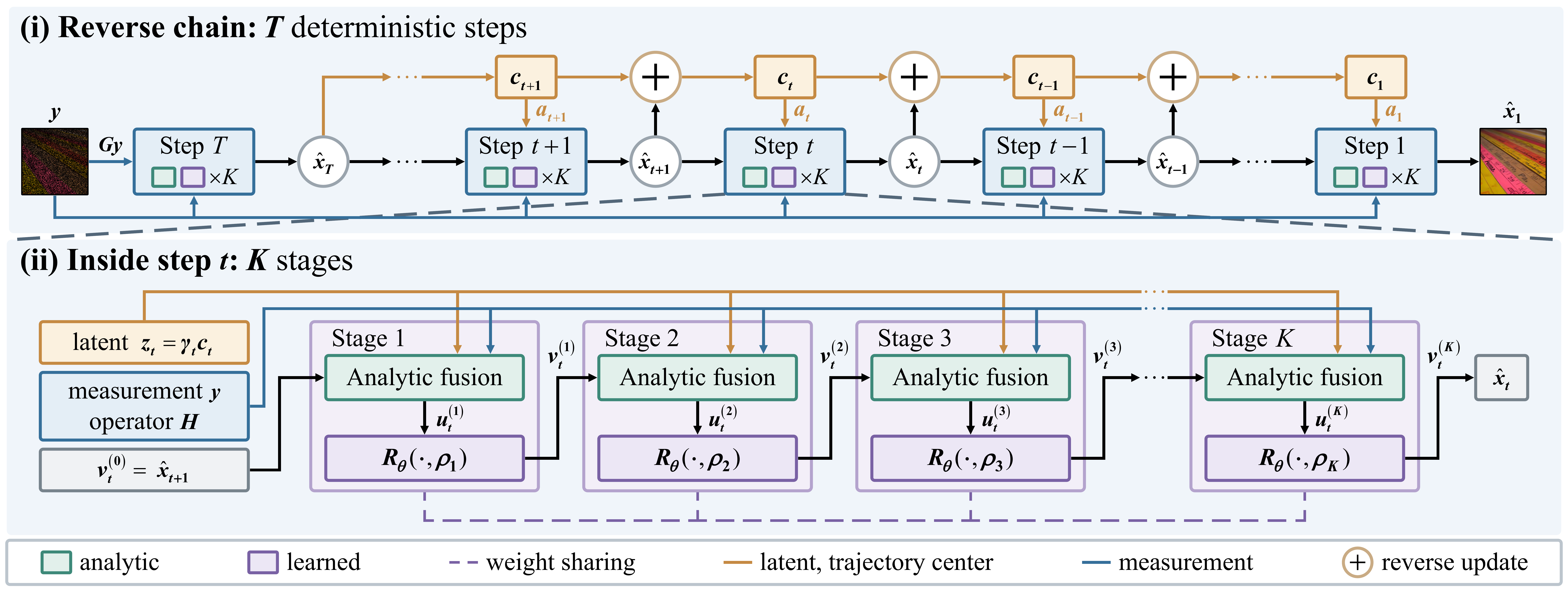}
\caption{\textbf{DORS.}
Training and inference run the same deterministic $T$-step chain from $\vx_T=\mG\vy$.
Each reconstruction $\xh_t$ initializes the next step and updates its propagated state through \cref{eq:reverse-update}.
For $t<T$, the trajectory center is fixed across the $K$ inner stages and enters fusion with weight $a_t$, and the initial step has $a_T=0$.
One refiner serves all steps and stages, receiving the fused image $\vu^{(k)}$ and $\rho_k$.}
\label{fig:framework}
\vspace{-5pt}
\end{figure*}

DORS alternates analytic fusion with a shared 3.7M-parameter refiner over $T$ deterministic reverse steps, each with $K$ inner stages (Fig.~\ref{fig:framework}). Each step's reconstruction initializes the next step and updates its trajectory center, which remains fixed during the inner stages. The measurement model enters through initialization and fusion, preserving the refiner interface across operators.
\subsection{Residual diffusion path}
\label{sec:path}

Let $\mG$ be a fixed lifting operator from measurements to the image grid: the identity for deblurring and inpainting, bicubic upsampling for super-resolution, and the zero-filled adjoint for MRI. We use the conditional Gaussian path
\begin{equation}\label{eq:path}
\begin{aligned}
q_{\mathrm{fwd}}(\vx_t\mid\vx_0,\vy)
&=\mathcal{N}\big(\vm_t(\vx_0,\vy),\tau_t^2\mI\big),\\
\vm_t(\vx,\vy)
&=\gamma_t\vx+(1-\gamma_t)\mG\vy,
\end{aligned}
\end{equation}
for $t=1,\dots,T$, with $T\ge2$, $s>0$, and
\begin{equation}\label{eq:schedule}
\gamma_t=\cos\Big(\frac{\pi t}{2T}\Big),
\qquad
\tau_t=s\sin\Big(\frac{\pi t}{2T}\Big).
\end{equation}
Thus $\tau_t^2=s^2(1-\gamma_t^2)$, with endpoints $(\gamma_0,\tau_0)=(1,0)$ and $(\gamma_T,\tau_T)=(0,s)$. The mean moves from the clean image to $\mG\vy$, as in residual and endpoint constructions~\cite{delbracio2023indi,liu2023i2sb,yue2023resshift}. The measurement operator $\mH$ and noise level $\sigma_y>0$ are known, and we use $s=1$ throughout.

Removing the known shift,
\begin{equation}\label{eq:shift}
\vz_t=\vx_t-(1-\gamma_t)\mG\vy,
\end{equation}
gives $\vz_t=\gamma_t\vx_0+\tau_t\veps$ under \cref{eq:path}, where $\veps\sim\mathcal{N}(\vzero,\mI)$ is independent of the image and measurement noise. This forward model specifies the Gaussian observation terms below. The executed chain starts from $\vz_T=\vzero$ and follows the deterministic recursion in Sec.~\ref{sec:chain}.

\subsection{Gaussian observation terms}
\label{sec:posterior}

Let $p_0(\vx)\propto\exp[-\Phi(\vx)]$ be the image prior. Together with \cref{eq:measurement}, the path defines
\begin{equation}\label{eq:joint}
q_{\mathrm{fwd}}(\vx_0,\vy,\vx_t)
=
p_0(\vx_0)\,p(\vy\mid\vx_0)\,
q_{\mathrm{fwd}}(\vx_t\mid\vx_0,\vy).
\end{equation}
Applying Bayes' rule gives the standard factorization.

\begin{proposition}[Forward-model posterior factorization]
\label{prop:posterior}
For $1\le t\le T$, the posterior under \cref{eq:joint} satisfies
\begin{equation}\label{eq:posterior}
\begin{aligned}
-\log q_{\mathrm{fwd}}(\vx\mid\vx_t,\vy)
&=\underbrace{\tfrac{1}{2\tau_t^2}\|\vz_t-\gamma_t\vx\|^2}_{\ldiff(\vx)}+\Phi(\vx)\\
&\quad+\underbrace{\tfrac{1}{2\sigma_y^2}\|\mH\vx-\vy\|^2}_{\lphys(\vx)}+\mathrm{const}.
\end{aligned}
\end{equation}
\end{proposition}
The two observation terms have precisions $a_t=\gamma_t^2/\tau_t^2$ and $\beta_y=\sigma_y^{-2}$. For $1\le t<T$, define $\vc_t=\vz_t/\gamma_t$, so that $\ldiff(\vu)=\tfrac{a_t}{2}\|\vu-\vc_t\|^2$. At $t=T$, $a_T=0$ and $\ldiff$ is constant, so the first step requires no center. DORS uses these quadratics along its deterministic chain. Sec.~\ref{sec:trajectory analysis} derives the center as a weighted combination of earlier reconstructions.

\subsection{Analytic precision fusion}
\label{sec:fusion}

Within step $t$, we hold the propagated state fixed and alternate fusion with refinement. We omit $t$ from the inner iterates. Given $\vv^{(k-1)}$ and $\rho_k>0$, fusion solves
\begin{equation}\label{eq:fusion}
\vu_\star^{(k)}
=
\operatorname*{arg\,min}_{\vu}\,
\ldiff(\vu)+\lphys(\vu)
+\tfrac{\rho_k}{2}\|\vu-\vv^{(k-1)}\|^2.
\end{equation}
The unique solution is
\begin{equation}\label{eq:fusion-solution}
\begin{aligned}
\mA_{t,k}
&=(a_t+\rho_k)\mI+\beta_y\mH^*\mH,\\
\vu_\star^{(k)}
&=\mA_{t,k}^{-1}
\Big[
\frac{\gamma_t}{\tau_t^2}\vz_t
+\beta_y\mH^*\vy
+\rho_k\vv^{(k-1)}
\Big].
\end{aligned}
\end{equation}
Where $\mH^*$ is the transpose for real operators and the Hermitian adjoint for complex MRI operators. 
After the residual shift in \cref{eq:shift}, this update has the same form as the joint-likelihood proximal step in UD$^{2}$M~\cite{mbakam2025ud2m}. 
DORS applies the update to the state propagated by its deterministic reconstruction chain. The numerical update $\vu^{(k)}$ equals $\vu_\star^{(k)}$ when the system is solved exactly, while super-resolution and MRI use warm-started conjugate gradients (CG).

For the exact solve, or CG initialized at $\vv^{(k-1)}$ in exact arithmetic, the observation terms satisfy
\begin{equation}\label{eq:descent}
\ldiff(\vu^{(k)})+\lphys(\vu^{(k)})
\le
\ldiff(\vv^{(k-1)})+\lphys(\vv^{(k-1)}).
\end{equation}
This is a single-fusion property at fixed $t$.

\subsection{Compact shared refinement}
\label{sec:refine}

Half-quadratic splitting~\cite{zhang2021dpir} pairs \cref{eq:fusion} with a prior update,
\begin{equation}\label{eq:prior-update}
\vv^{(k)}
=
\operatorname*{arg\,min}_{\vv}\,
\Phi(\vv)+\tfrac{\rho_k}{2}\|\vv-\vu^{(k)}\|^2.
\end{equation}
DORS replaces this step with a shared learned image update,
\begin{equation}\label{eq:refine}
\vv^{(k)}=R_\theta\big(\vu^{(k)},\rho_k\big).
\end{equation}
We use a 3.7M-parameter NAFNet-style network~\cite{chen2022nafnet}, encoding the coefficient through a constant channel $\sigma_{\mathrm{prior},k}=\rho_k^{-1/2}$, as in noise-level-conditioned denoisers~\cite{zhang2018ffdnet,zhang2021dpir}. The same weights serve all $t$ and $k$. We train $R_\theta$ from scratch as an image update through reconstruction supervision on the complete chain.

The refiner receives only $(\vu^{(k)},\rho_k)$, with no separate operator, measurement, state, or timestep input. A new measurement model is incorporated through the lifting $\mG$ and the fusion system while retaining this input interface. Sec.~\ref{sec:transfer_dors} evaluates transfer across operators and tasks with fixed refiner weights.

\paragraph{Coefficient schedule and warm start.}
For $K>1$, the coefficients are log-linear,
\begin{equation}\label{eq:rho}
\log\rho_k
=
\log\rho_{\min}
+\frac{k-1}{K-1}
\big(\log\rho_{\max}-\log\rho_{\min}\big),
\end{equation}
with positive endpoints learned during training. For $K=1$, $\rho_1=\sqrt{\rho_{\min}\rho_{\max}}$.
The first step starts from $\vv_T^{(0)}=\mG\vy$, and later steps use $\vv_t^{(0)}=\xh_{t+1}$.
Each step returns $\xh_t=\vv_t^{(K)}$, with final reconstruction $\xh_1$.
The default $T=K=4$ uses 16 refiner evaluations with one set of shared weights.

\begin{figure*}[!t]
\centering
\includegraphics[width=1\textwidth]{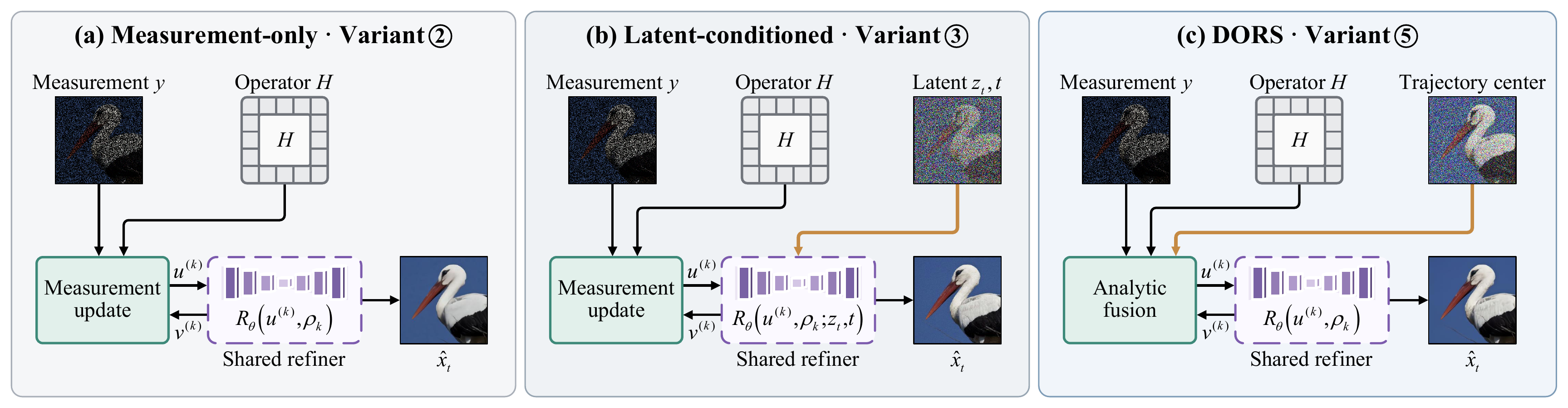}
\vspace{-15pt}
\caption{\textbf{Mechanism Analysis}
(a) Analytic measurement updates with shared refinement.
(b) Additional state and timestep inputs to the refiner.
(c) DORS uses the trajectory center in analytic fusion.
Its refiner takes only the fused estimate and a scalar
proximal coefficient.
The panels correspond to variants \ding{173}, \ding{174},
and \ding{176} in Tab.~\ref{tab:controlled}.}
\label{fig:factorization}
\vspace{-6pt}
\end{figure*}

\subsection{Deterministic reverse recursion and training}
\label{sec:chain}

We realize \cref{eq:path} with a Gaussian Markov coupling whose reverse conditional mean is affine in $\vx_0$. Substituting $\xh_t$ for $\vx_0$ in this mean and removing the shift gives,
\begin{equation}\label{eq:reverse-update}
\begin{aligned}
\vz_{t-1}
&=\kappa_t\vz_t+
(\gamma_{t-1}-\kappa_t\gamma_t)\,\xh_t,\\
\kappa_t
&=\frac{\gamma_t}{\gamma_{t-1}}
\frac{\tau_{t-1}^2}{\tau_t^2}, \quad t=T,\dots,2 .
al\end{aligned}
\end{equation}
This coupling specifies $\kappa_t$, and the marginals in \cref{eq:path} alone leave the coupling open. Starting from $\vz_T=\vzero$, the recursion draws no noise and defines a map from $(\vy,\mH)$ to $\xh_1$.

\paragraph{Full-chain reconstruction training.}
Training unrolls the same chain used at inference and differentiates through every fusion, refinement, warm start, and reverse update. For each training pair and its operator, we minimize $\mathcal{L}=T^{-1}\sum_{t=1}^{T}\mathcal{L}_t$, where
\begin{equation}\label{eq:loss}
\mathcal{L}_t
=
\|\xh_t-\vx_0\|_2^2
+\lambda_{\mathrm{aux}}
\sum_{k=1}^{K-1}w_k\|\vv_t^{(k)}-\vx_0\|_2^2,
\end{equation}
with $\lambda_{\mathrm{aux}}=0.05$ and $w_k=k/K$.
Supervision acts on the states visited by the solver, and earlier reconstructions receive gradients through later steps that use them.
This objective trains DORS as a deterministic estimator, optimizing $\theta$ and the two endpoints of the proximal-coefficient schedule.

\begin{figure*}[!t]
\centering
\includegraphics[width=1\textwidth]{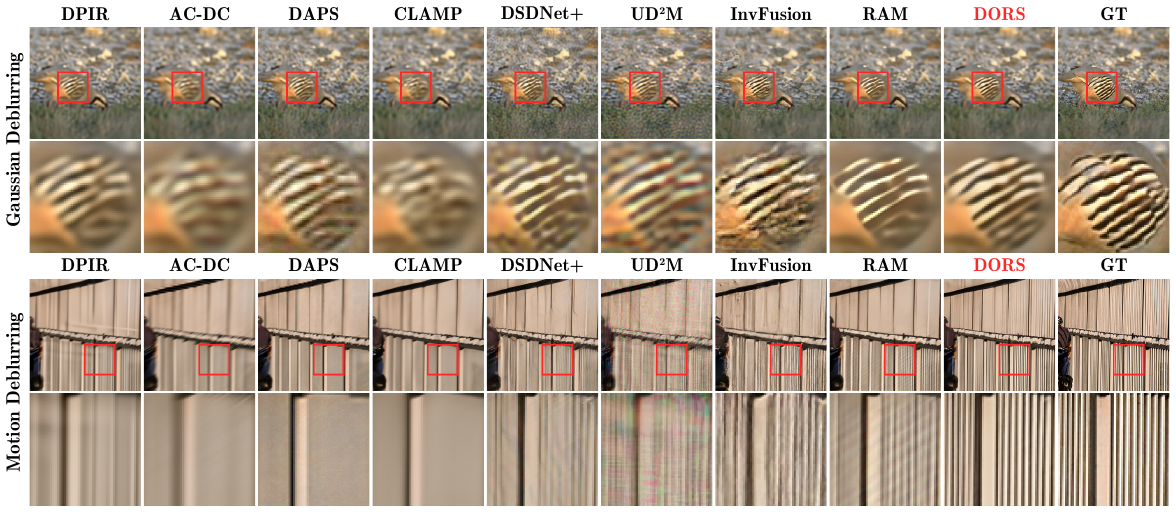}
\caption{\textbf{Visualization of deblurring ($\sigma_y=0.05$)}. Top: Gaussian kernel. Bottom: motion kernel.}
\vspace{-5pt}
\label{fig:deblurring-figure}
\end{figure*}

\subsection{Trajectory center and its analytic effect}
\label{sec:trajectory analysis}

The state propagated by \cref{eq:reverse-update} has an exact representation in terms of earlier reconstructions. The centers $\vc_t=\vz_t/\gamma_t$ are defined for $1\le t<T$. The final reconstruction is $\hat{\vx}_1=\vv_1^{(K)}$.

\begin{proposition}[Trajectory center]
\label{prop:center}
For any reconstructions $\xh_t$ used in \cref{eq:reverse-update}, the first center is $\vc_{T-1}=\xh_T$. For $2\le t<T$,
\begin{equation}\label{eq:center-recursion}
\vc_{t-1}
=
\alpha_t\vc_t+(1-\alpha_t)\xh_t,
\qquad
\alpha_t=\frac{a_t}{a_{t-1}}.
\end{equation}
Consequently, for $1\le t<T$,
\begin{equation}\label{eq:center-history}
\vc_t
=
\sum_{j=t+1}^{T}
\frac{a_{j-1}-a_j}{a_t}\,\xh_j.
\end{equation}
The coefficients are nonnegative and sum to one.
\end{proposition}

The center weights are independent of $s$.
For the default $T=4$, $\alpha_2=\alpha_3=3-2\sqrt{2}\approx0.17$, so the two updates after initialization form an exponential moving average with constant decay.
This history representation holds for any reconstruction sequence under \cref{eq:reverse-update}. DORS places the resulting center in the analytic fusion.

To isolate the center's effect, consider one fusion at $t<T$ with the same input $\vv$, operator, and coefficients.
Replacing $\vc_t$ by $\vv$ combines the two penalties centered at $\vv$ into one with coefficient $\rho_k+a_t$, preserving $\mA_{t,k}$.
Let $\vu_{\mathrm{cur}}$ be its exact solution.
The difference from \cref{eq:fusion-solution} is
\begin{equation}\label{eq:center-difference}
\vu_\star-\vu_{\mathrm{cur}}
=
a_t\mA_{t,k}^{-1}(\vc_t-\vv).
\end{equation}
For an eigenvector $\bm{q}_i$ with $\mH^*\mH\bm{q}_i=\lambda_i\bm{q}_i$, this becomes
\begin{equation}\label{eq:center-spectrum}
\langle\bm{q}_i,\vu_\star-\vu_{\mathrm{cur}}\rangle
=
\frac{a_t}{a_t+\rho_k+\beta_y\lambda_i}
\langle\bm{q}_i,\vc_t-\vv\rangle.
\end{equation}
The response decreases with measurement strength $\lambda_i$: the analytic update attenuates center corrections more strongly in well-observed directions.
For the exact fusion, this coefficient is $a_t/(a_t+\rho_k)$ on $\operatorname{null}(\mH)$. Both the measurement-only ($a_t=0$) and current-center fusions preserve the input iterate's component in this subspace.
Sec.~\ref{sec:controlled} compares DORS with these controls and latent-conditioned refinement at matched compute.
Further derivations and details are provided in the Appendix.

% % ==================================================================
\section{Experiments}\label{sec:experiments}

% ==================================================================
\subsection{Setup and Experimental Settings}\label{sec:fairsettings}

\paragraph{DORS models and evaluation settings.}
We train separate deblurring, inpainting, and $\times4$ SR
models on $256\times256$ DIV2K/Flickr2K \citep{agustsson2017ntire,timofte2017ntire} crops at
$\sigma_y=0.05$, using $31\times31$ blur kernels, random
masks removing 30--70\% of pixels, and bicubic downsampling,
respectively.
Standard evaluation uses $256\times256$ BSD100 \citep{martin2001bsd} center crops
with matching degradation settings (Sec.~\ref{sec:evaluation}).
For fastMRI brain ~\citep{zbontar2018fastmri}, one single-coil model handles both
$\times4$ and $\times8$ acceleration with $\mH=MF$,
random Cartesian masks, and $\sigma_y=0.01$.
Transfer reuses the deblurring or single-coil model with
fixed refiner weights and $T=K=4$. Transferring evaluated target settings and
inference details are specified in
Sec.~\ref{sec:transfer_dors}.

\paragraph{Implementation.}
Each model uses a shared 3.7M-parameter refiner with
$T=K=4$ and $s=1$, and is trained from scratch for
300k iterations using AdamW~\citep{loshchilov2019adamw} on one NVIDIA RTX PRO 6000 GPU with $(\beta_1,\beta_2)=(0.9,0.999)$, weight decay $10^{-3}$, and a learning rate of $2\times10^{-4}$. The module is a three-level NAFNet-style U-Net
of width 64 with 14 NAFBlocks.
Training and inference use the same deterministic chain.

\paragraph{Compared Baselines and Metrics.}
During comparison in Sec.~\ref{sec:evaluation}, zero-shot and plug-and-play methods (DPIR, AC-DC, DAPS, and CLAMP) use released pretrained priors. RAM uses the released unified model supplied with the test operator.
i-DEQ in the inpainting task uses its official weights.
DSDNet+ and UD$^2$M use their released weights, following the official instruction. They run with the test operator: DSDNet+ takes the blur kernel as a network input, and for UD$^2$M (official released weight), the test kernel or mask replaces the training operator in its closed-form data-consistency step, except in super-resolution, where it keeps its operators. For those settings different from DORS, we retrain ($^*$) InvFusion (single-step MMSE output), OFTSR, PDNet, uDPIR (the unrolled DPIR baseline~\cite{terris2026ram}, untied), and single-coil
MoDL on the corresponding DORS source-task data, operators, and noise levels to keep fair comparison. (\textendash): result not reported, or the method does not support the task.
We report PSNR and SSIM, with Q-Align~\cite{wu2023q} and
VisualQuality-R1 (VQ-R1)~\cite{wu2026visualquality} scores additionally
reported for perceptual quality.

\begin{figure*}[!t]
\centering
\includegraphics[width=1\textwidth]{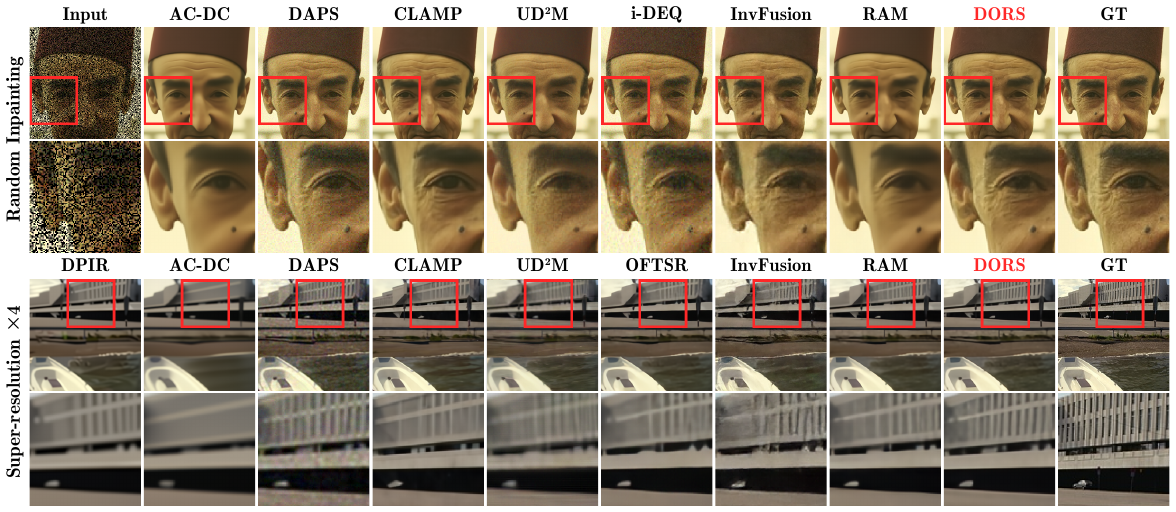}
\caption{\textbf{Visualization of inpainting and $\times4$ super-resolution ($\sigma_y=0.05$)}. Top: random-mask inpainting. Bottom: super-resolution. }
\label{fig:srrring}
\vspace{-3pt}
\end{figure*}

\begin{table*}[!tbp]
\caption{\textbf{Matched-compute controlled study.}
All variants are trained with the same backbone and training schedule, and differ only in how measurement $\ell_{\mathrm{phys}}$ and latent information $\ell_{\mathrm{diff}}$ are used.
Tests use BSD100 at $\sigma_y=0.05$, and fastMRI brain at $\sigma_y=0.01$. \best{Red}: best
}
\label{tab:controlled}
\centering
\footnotesize
\setlength{\tabcolsep}{2pt}
\renewcommand{\arraystretch}{1.04}
\begin{tabular*}{\linewidth}{@{\extracolsep{\fill}}>{\raggedright\arraybackslash}p{0.26\linewidth}c>{\raggedright\arraybackslash}p{0.17\linewidth}*{2}{cc}@{}}
\toprule
\multirow{2}{*}{\textbf{Variant}} & \multirow{2}{*}{$\lphys$} & \multirow{2}{*}{$\ldiff$ \textbf{information}} & \multicolumn{2}{c}{\textbf{Deblurring (Motion)}} & \multicolumn{2}{c}{\textbf{MRI (\textbf{Single-coil}$\times4$)}} \\
\cmidrule(lr){4-5}\cmidrule(lr){6-7}
 & & & \textbf{PSNR}$\uparrow$ & \textbf{SSIM}$\uparrow$ & \textbf{PSNR}$\uparrow$ & \textbf{SSIM}$\uparrow$ \\
\midrule
\ding{172} Conditional denoiser & learned &learned & 23.64 & 0.603 & 28.85 & 0.637 \\
\ding{173} DORS w/o diffusion likelihood & analytic & none & 25.08 & 0.675 & 29.78 & 0.650 \\
\ding{174} Phys.-analytic + diff.-conditioned & analytic & learned, in refinement & 25.28 & 0.681 & 30.03 & 0.660 \\
\ding{175} Curvature-matched current-center & analytic & precision $a_t$, no center & 24.95 & 0.674 & 29.40 & 0.645 \\
% \ding{176}  &  &  & 25.31 & 0.682 & 30.01 & 0.656 \\
 \midrule
\ding{176} \ourrow{} & analytic & analytic, center $\vc_t$ & \best{25.75} & \best{0.705} & \best{30.47} & \best{0.668} \\
\bottomrule

\end{tabular*}
\vspace{-5pt}
\end{table*}

\begin{table}[t]

\caption{\textbf{Paired differences between controlled variants in Tab.~\ref{tab:controlled}.}
Entries are mean differences (first minus second) with 95\% bootstrap confidence intervals below. Positive values favor the first.}
\label{tab:controlled_ci}
\centering
\footnotesize
\setlength{\tabcolsep}{2pt}
\renewcommand{\arraystretch}{1.04}
\begin{tabular*}{\columnwidth}{@{\extracolsep{\fill}}lcccc@{}}
\toprule
\multirow{2}{*}{\textbf{Comparison}} & \multicolumn{2}{c}{\textbf{Deblurring}} & \multicolumn{2}{c}{\textbf{MRI} ($\times4$)} \\
\cmidrule(lr){2-3}\cmidrule(l){4-5}
 & $\Delta$PSNR & $\Delta$SSIM & $\Delta$PSNR & $\Delta$SSIM \\
\midrule
\multirow{2}{*}{\ding{176}$-$\ding{174}} & +0.47 & +0.024 & +0.44 & +0.008 \\[-1pt]
 & \dci{0.42}{0.53} & \dci{.021}{.027} & \dci{0.43}{0.45} & \dci{.007}{.010} \\
\addlinespace[3pt]
\multirow{2}{*}{\ding{176}$-$\ding{175}} & +0.80 & +0.031 & +1.07 & +0.024 \\[-1pt]
 & \dci{0.71}{0.90} & \dci{.027}{.034} & \dci{1.05}{1.09} & \dci{.022}{.025} \\
\addlinespace[3pt]
\multirow{2}{*}{\ding{176}$-$\ding{173}} & +0.67 & +0.029 & +0.69 & +0.019 \\[-1pt]
 & \dci{0.59}{0.74} & \dci{.027}{.032} & \dci{0.68}{0.71} & \dci{.017}{.020} \\
\addlinespace[3pt]
\multirow{2}{*}{\ding{173}$-$\ding{175}} & +0.13 & +0.001 & +0.38 & +0.005 \\[-1pt]
 & \dci{0.11}{0.16} & \dci{.000}{.002} & \dci{0.37}{0.38} & \dci{.005}{.005} \\
\bottomrule
\end{tabular*}
\vspace{-6pt}
\end{table}

\subsection{Mechanism Ablations }\label{sec:controlled}

We test how a compact deterministic solver uses its propagated state and whether center placement contributes beyond additional quadratic curvature.
All variants in Tab.~\ref{tab:controlled} are trained separately with the same backbone, training data, and full-chain deterministic training schedule.
We fix $T=K=4$ and match the number of training iterations.

In Tab.~\ref{tab:controlled}, the \emph{\ding{172} conditional denoiser} combines the latent and lifted measurement $\mG\vy$ inside a network conditioned on the timestep. \emph{\ding{173} DORS w/o diffusion likelihood} retains the analytic measurement update and the shared refiner but removes the latent term, yielding an unrolled HQS baseline with the same number of parameters and computational cost, in the same category as USRNet~\cite{zhang2020usrnet}. \emph{\ding{174} Phys.-analytic + diff.-conditioned} uses the same measurement-only fusion and supplies $\vz_t$ and $t$ as additional refiner inputs. This variant is inspired by Diff-Unfolding's public design.
The \emph{\ding{175} curvature-matched current-center} control retains the additional precision $a_t$ and replaces the trajectory center $\vc_t$ with the current iterate $\vv$ in every fusion. DORS uses both observation terms analytically, centers the latent quadratic at $\vc_t$ and holds this propagated center fixed across the inner stages of each reverse step. Tab.~\ref{tab:controlled_ci} reports mean paired differences and 95\% bootstrap confidence intervals, resampling images for deblurring and volumes for MRI.

\paragraph{Analytic use of the propagated state.}
All variants with an analytic measurement update outperform \ding{172}, and supplying the state to the refiner (\ding{174}) improves over omitting the latent term (\ding{173}).
DORS achieves the highest PSNR and SSIM on both tasks, exceeding \ding{174} by 0.47\,dB on motion deblurring and 0.44\,dB on MRI.
In Tab.~\ref{tab:controlled_ci}, the paired confidence intervals are positive for metrics on both tasks.
These results support analytic use of the propagated state in the compact solver.

\paragraph{Value of the trajectory center.}
DORS improves over the current-center control on deblurring and  MRI, again with positive confidence intervals for PSNR and SSIM. It also improves over removing the latent term. 
These comparisons show that using the trajectory center in the fusion improves reconstruction beyond adding quadratic curvature alone. More ablation studies are reported in the Appendix.

\begin{table*}[!tbp]
\caption{\textbf{Non-blind deblurring (Gaussian and motion kernels) and random-mask inpainting on BSD100.} 
\best{Red}: best, \second{Blue}: second-best.}
\label{tab:deblur}
\centering
\footnotesize
\setlength{\tabcolsep}{2pt}
\renewcommand{\arraystretch}{1.04}
\begin{tabular*}{\linewidth}{@{\extracolsep{\fill}}lc*{3}{cccc}@{}}
\toprule
\multirow{2}{*}{\textbf{Method}} &\multirow{2}{*}{\textbf{Type}}& \multicolumn{4}{c}{\textbf{Deblur (Gaussian)}} & \multicolumn{4}{c}{\textbf{Deblur (Motion)}} & \multicolumn{4}{c}{\textbf{Inpaint (Random)}} \\
\cmidrule(lr){3-6}\cmidrule(lr){7-10}\cmidrule(lr){11-14}
 &  & \textbf{PSNR$\uparrow$} & \textbf{SSIM$\uparrow$} & \textbf{Q-Align$\uparrow$} & \textbf{VQ-R1$\uparrow$} & \textbf{PSNR$\uparrow$} & \textbf{SSIM$\uparrow$} & \textbf{Q-Align$\uparrow$} & \textbf{VQ-R1$\uparrow$} & \textbf{PSNR$\uparrow$} & \textbf{SSIM$\uparrow$} & \textbf{Q-Align$\uparrow$} & \textbf{VQ-R1$\uparrow$} \\
\midrule
DPIR \citep{zhang2021dpir} & zero-shot & 25.39 & 0.689 & 1.607 & 1.634 & 25.04 & 0.667 & 1.871 & 1.594 & \textendash & \textendash & \textendash & \textendash \\
%PnP-Flow & Plug-and-play \\
AC-DC \citep{shrestha2026taming} & zero-shot & 24.64 & 0.645 & 1.418 & 1.351 & 23.71 & 0.599 & 1.317 & 1.117 & 26.29 & 0.717 & 2.649 & 2.789 \\
%DDiff$^\S$ \citep{Kim_2026_CVPR} & Plug-and-play & 17.55 & 0.252 & 1.113 & 1.000 & 14.83 & 0.184 & 1.022 & 1.000 & 27.08 & 0.759 & 2.597 & 2.808 \\
DAPS \citep{zhang2025daps} & zero-shot & 24.06 & 0.578 & 1.294 & 1.155 & 24.03 & 0.597 & 1.630 & 1.320 & 24.74 & 0.614 & 2.244 & 2.572 \\
CLAMP \citep{pmlr-v306-shin26b} & zero-shot & 25.25 & 0.672 & 1.845 & 1.866 & 24.14 & 0.627 & 1.865 & 1.688 & \second{28.33} & \second{0.844} & \best{2.938} & \second{3.252} \\
DSDNet+ \citep{kuo2025efficient} & supervised & 25.05 & 0.661 & 1.781 & 1.818 & 24.63 & 0.663 & 1.878 & 1.478 & \textendash & \textendash & \textendash & \textendash \\
UD$^2$M  \citep{mbakam2025ud2m} & supervised & 24.25 & 0.611 & 1.491 & 1.397 & 23.20 & 0.558 & 1.417 & 1.100 & 26.20 & 0.724 & 2.307 & 2.595 \\
i-DEQ \citep{clerc2026ideq} & supervised & \textendash & \textendash & \textendash & \textendash & \textendash & \textendash & \textendash & \textendash & 26.22 & 0.713 & 1.986 & 2.191 \\
{InvFusion}$^*$ \citep{elata2025invfussion} & supervised & \second{25.78} & 0.696 & 1.971 & 1.673 & 24.45 & 0.632 & 1.835 & 1.246 & 28.28 & 0.829 & 2.359 & 2.765 \\
RAM \citep{terris2026ram} & supervised & 25.40 & \second{0.699} & \second{2.033} & \second{1.897} & \second{25.06} & \second{0.683} & \second{2.017} & \second{1.823} & 27.71 & 0.789 & 2.656 & 2.910 \\
\midrule
\textbf{DORS} (ours) & supervised & \best{26.00} & \best{0.706} & \best{2.059} & \best{1.924} & \best{25.75} & \best{0.705} & \best{2.115} & \best{2.017} & \best{29.31} & \best{0.863} & \second{2.664} & \best{3.294} \\
\bottomrule
\end{tabular*}
\end{table*}

\begin{table}[!tbp]
\caption{\textbf{$\times4$ SR on BSD100 and ImageNet.} (a)~On BSD100 ($\sigma_{y}=0.05$), DORS is trained for SR.
(b)~Operator transfer on ImageNet ($\sigma_{y}=0.025$): the deblurring DORS model of Tab.~\ref{tab:deblur}, with $H$ swapped for SR and no retraining.
 Compared baselines use the same SR weights in (a) and (b).}
\label{tab:sr_imag}
\centering
\footnotesize
\setlength{\tabcolsep}{2pt}
\renewcommand{\arraystretch}{1.04}
\begin{tabular*}{\linewidth}{@{\extracolsep{\fill}}lc*{2}{cc}@{}}
\toprule
 & & \multicolumn{2}{c}{\textbf{(a) BSD100}} & \multicolumn{2}{c}{\textbf{(b) ImageNet}} \\
\cmidrule(lr){3-4}\cmidrule(lr){5-6}
\textbf{Method} & \textbf{Type} & \textbf{PSNR$\uparrow$} & \textbf{SSIM$\uparrow$} & \textbf{PSNR$\uparrow$} & \textbf{SSIM$\uparrow$} \\
\midrule
DPIR \citep{zhang2021dpir} & zero-shot & 23.84 & 0.613 & 26.69 & 0.738 \\
AC-DC \citep{shrestha2026taming} & zero-shot & 22.86 & 0.541 & 26.31 & 0.720 \\
DAPS \citep{zhang2025daps} & zero-shot & 23.03 & 0.537 & 25.52 & 0.639 \\
CLAMP \citep{pmlr-v306-shin26b} & zero-shot & 23.86 & 0.615 & \second{26.97} & \second{0.745} \\
UD$^2$M \citep{mbakam2025ud2m} & supervised & 22.60 & 0.579 & 24.61 & 0.701 \\
OFTSR$^*$ \citep{zhu2026onestep} & supervised & \second{24.19} & \second{0.623} & 26.44 & 0.710 \\
{InvFusion}$^*$ \citep{elata2025invfussion} & supervised & 24.13 & 0.611 & 26.31 & 0.693 \\
RAM \citep{terris2026ram} & supervised & 24.05 & 0.617 & 26.65 & 0.737 \\
\midrule
\ourrow{} & supervised & \best{24.35} & \best{0.631} & \best{27.08} & \best{0.747} \\
\bottomrule
\end{tabular*}
\end{table}

\begin{table*}[!tbp]
\caption{\textbf{Brain MRI reconstruction on fastMRI.} (a)~Single-coil: DORS is trained for this task. (b)~Multi-coil: the single-coil DORS model is transferred to the multi-coil operator, no retraining. Evaluation is magnitude-based over the whole image. $^\diamond$: inference hyperparameters tuned on training slices. MoDL: retrained for single-coil, released checkpoint for multi-coil.}
\label{tab:mri}
\centering
\footnotesize
\setlength{\tabcolsep}{2pt}
\renewcommand{\arraystretch}{1.04}
\begin{tabular*}{\linewidth}{@{\extracolsep{\fill}}lc*{4}{c@{\hspace{4pt}}c}@{}}
\toprule
\multirow{2}{*}{\textbf{Method}} &\multirow{2}{*}{\textbf{Type}}& \multicolumn{2}{c}{\textbf{(a) Single-coil} \textbf{$\times4$}} & \multicolumn{2}{c}{\textbf{(a) Single-coil} \textbf{$\times8$}} & \multicolumn{2}{c}{\textbf{(b) Multi-coil} \textbf{$\times4$}} & \multicolumn{2}{c}{\textbf{(b) Multi-coil} \textbf{$\times8$}} \\
\cmidrule(lr){3-4}\cmidrule(lr){5-6}\cmidrule(lr){7-8}\cmidrule(lr){9-10}
 &  & \textbf{PSNR$\uparrow$} & \textbf{SSIM$\uparrow$} & \textbf{PSNR$\uparrow$} & \textbf{SSIM$\uparrow$} & \textbf{PSNR$\uparrow$} & \textbf{SSIM$\uparrow$} & \textbf{PSNR$\uparrow$} & \textbf{SSIM$\uparrow$} \\
\midrule
CG-SENSE \citep{pruessmann2001cgsense} & classic & \textendash & \textendash & \textendash & \textendash & 30.14 & \second{0.818} & 26.56 & 0.768 \\
DPIR$^\diamond$ \citep{zhang2021dpir} & zero-shot & 29.18 & 0.576 & 25.25 & 0.444 & \second{36.48} & \best{0.898} & \second{31.94} & \second{0.846} \\
PDNet$^*$ \citep{adler2018pdnet} & supervised & 29.04 & 0.639 & 25.40 & 0.508 & 30.07 & 0.687 & 25.58 & 0.606 \\
uDPIR$^*$ \citep{terris2026ram} & supervised & \second{30.04} & 0.658 & \second{27.75} & 0.549 & -8.96 & 0.149 & -9.98 & 0.107 \\
MoDL$^*$ \citep{aggarwal2019modl} & supervised & 29.38 & \best{0.671} & 25.57 & \second{0.552} & 33.23 & 0.768 & 27.78 & 0.690 \\
RAM \citep{terris2026ram} & supervised & 28.96 & 0.616 & 27.18 & 0.496 & 22.68 & 0.439 & 21.87 & 0.384 \\
\midrule
\ourrow{} & supervised & \best{30.47} & \second{0.668} & \best{28.70} & \best{0.571} & \best{36.75} & \best{0.898} & \best{33.47} & \best{0.860} \\
\bottomrule
\end{tabular*}
\end{table*}

\subsection{Evaluation}\label{sec:evaluation}
\begin{figure*}[!t]
\centering
\vspace{-5pt}
\includegraphics[width=\textwidth]{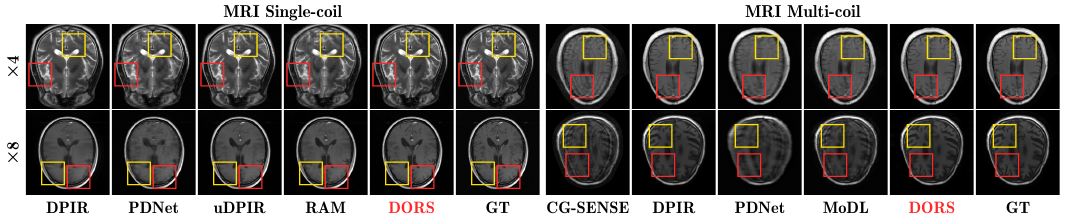}
\caption{\textbf{Visualization of MRI reconstruction at $\times4$ and $\times8$ acceleration.} Left: single-coil. Right: multi-coil, where DORS is the single-coil model applied with the multi-coil operator and no retraining. Boxes mark the regions where the methods differ.}
\vspace{-4pt}
\label{fig:mriiing}
\end{figure*}

We evaluate the task-specific DORS models under the settings and compared baselines 
in Sec.~\ref{sec:fairsettings}. 

\subsubsection{Deblurring, Inpainting and Super-resolution}
Table~\ref{tab:deblur} compares Gaussian deblurring, motion
deblurring, and random-mask inpainting on $256\times256$ center crops of BSD100 \citep{martin2001bsd}.
DORS achieves the highest PSNR, SSIM, and VQ-R1 in all three
settings, and the highest Q-Align scores in both deblurring
settings.
The SR-trained DORS model reaches 24.35\,dB and 0.631 SSIM
in the (a) BSD100 columns of Tab.~\ref{tab:sr_imag}, the highest
values among the compared methods. Fig.~\ref{fig:deblurring-figure} and Fig.~\ref{fig:srrring} show that DORS restores sharper edges with fewer artifacts, preserving fine structures.
Further natural-image evaluations are reported in the Appendix.

\subsubsection{Single-coil MRI reconstruction}\label{sec:mri}

The single-coil model is trained on 70,748 fastMRI brain
slices and evaluated on 21,842 validation slices.
DPIR applies its released denoiser separately to the real
and imaginary parts, with inference parameters tuned on
training slices ($\diamond$).
In Tab.~\ref{tab:mri} (a), DORS
achieves the highest PSNR at both accelerations, reaching
30.47\,dB at $\times4$ and 28.70\,dB at $\times8$.
It also achieves the highest SSIM at $\times8$. Visualization in Fig.~\ref{fig:mriiing} shows that DORS recovers more brain contours and clearer internal structures. The references carry background noise, and the Appendix reports evaluation inside the brain.

\begin{table*}[t]
\centering
\footnotesize
\setlength{\tabcolsep}{2pt}
\renewcommand{\arraystretch}{1.04}
\begin{minipage}[t]{\columnwidth}
\centering
\caption{\textbf{Generalization evaluation on ImageNet (Gaussian deblurring).}
DORS uses its original models without retraining. All settings, including randomly selected test data, follow DAPS~\citep{zhang2025daps}.
}
\label{tab:imagenet}
\begin{tabular*}{\linewidth}{@{\extracolsep{\fill}}lccccc@{}}
\toprule
\textbf{Method} & \textbf{Type} & \textbf{PSNR$\uparrow$} & \textbf{SSIM$\uparrow$} & \textbf{Q-Align$\uparrow$} & \textbf{VQ-R1$\uparrow$} \\
\midrule
DPIR \citep{zhang2021dpir} & zero-shot & 26.39 & 0.727 & 1.409 & 1.302 \\
%PnP-Flow & Plug-and-play \\
AC-DC \citep{shrestha2026taming} & zero-shot & 25.93 & 0.698 & 1.339 & 1.227 \\
%DDiff \citep{Kim_2026_CVPR} & Plug-and-play & 23.06 & 0.532 & 1.920 & 1.642 \\
DAPS \citep{zhang2025daps} & zero-shot & 25.92 & 0.658 & 1.408 & 1.237 \\
CLAMP \citep{pmlr-v306-shin26b} & zero-shot & 26.36 & 0.712 & 1.831 & \best{1.641} \\
%DMBSR \citep{laroche2023dmbsr} & supervised, unrolled & 26.95 & 0.743 & 2.120 & 1.829 \\
%DeepSN-Net$^*$ & supervised & 26.26 & 0.695 & 1.698 & 1.141 \\
DSDNet+ \citep{kuo2025efficient} & supervised & 24.79 & 0.636 & 1.124 & 1.070 \\
UD$^2$M  \citep{mbakam2025ud2m} & supervised & 24.63 & 0.613 & 1.063 & 1.072 \\
InvFusion$^*$ \citep{elata2025invfussion} & supervised & \second{26.96} & 0.726 & 1.911 & 1.306 \\
RAM \citep{terris2026ram} & supervised & 25.96 & \second{0.735} & \best{1.973} & 1.523 \\
\midrule
\ourrow{} & supervised & \best{27.40} & \best{0.748} & \second{1.933} & \second{1.582} \\
% DORS ($T{=}2$) & dual-observation & 27.35 & 0.747 & 1.888 & 1.508 \\
\bottomrule
\end{tabular*}
\end{minipage}
\hfill\hfill
%---------------- left: efficiency ----------------
\begin{minipage}[t]{\columnwidth}
\centering
\caption{\textbf{Inference cost on BSD100 deblurring}
at $256\times256$ with batch size 1.
Parameter counts include frozen pretrained networks. $^\dagger$: pretrained unconditional ImageNet prior. $^\ddagger$: mean over images.}
\label{tab:efficiency}
\begin{tabular*}{\linewidth}
{@{\extracolsep{\fill}}lccccc@{}}
\toprule
\textbf{Method} & \textbf{Params (M)}  & \textbf{FFTs} & \textbf{MACs (T)}  & \textbf{Mem.(GB)}  & \textbf{Time (s)}\\
     
\midrule
DPIR~\citep{zhang2021dpir}
& 32.6 & \textbf{18} & 1.11 & 0.219 & 0.109 \\
AC-DC~\citep{shrestha2026taming}
& 552.8$^\dagger$ & 93,450$^\ddagger$ & 2710.0 & 2.66 & 209.7 \\
DAPS~\citep{zhang2025daps}
& 552.8$^\dagger$ & 80,001 & 1119.8 & 2.67 & 87.3\\
CLAMP~\citep{pmlr-v306-shin26b}
& 552.8$^\dagger$ & 1,701 & 448.6 & 9.77 & 39.9 \\
DSDNet+~\citep{kuo2025efficient}
& 131.3 & \textbf{18} & 0.425 & 1.08 & 0.077 \\
UD$^2$M~\citep{mbakam2025ud2m}
& 129.2 & 672$^\ddagger$ & 1.91 & 0.791 & 0.265 \\
InvFusion$^*$~\citep{elata2025invfussion}
& 170.4 & 67 & \textbf{0.047} & 0.870 & \textbf{0.035} \\
RAM~\citep{terris2026ram}
& 35.6 & 635$^\ddagger$ & 0.166
& 0.358 & 0.039 \\
\midrule
\ourrow{}
& \textbf{3.7} & 35 & 0.408
& \textbf{0.157} & 0.075 \\
\bottomrule
\end{tabular*}
\end{minipage}

\vspace{-4pt}  
\end{table*}

\subsection{Transferability across Operators and Tasks}
\label{sec:transfer_dors}
DORS retains its refiner weights and $T=K=4$. Initialization and analytic fusion use the target $\mH$ and $\sigma_y$. Implementation details for DORS and baselines are in the Appendix.

\noindent\textbf{Deblurring under image and kernel shifts.}\label{sec:imagenet}
We reuse the Gaussian deblurring model of
Tab.~\ref{tab:deblur} on $256\times256$ ImageNet~\cite{deng2009imagenet} images
under the DAPS protocol~\citep{zhang2025daps}.
The target uses a $61\times61$ Gaussian kernel with standard
deviation 3 and differs from the BSD100 setting in
distribution and blur operator.
Compared baselines reuse corresponding deblurring checkpoints (zero-shot models and RAM use released unified model supplied in the deblurring task, and InvFusion retains its Tab.~\ref{tab:deblur} weights), while DSDNet+ and UD$^2$M apply pretrained priors with changing test operators substituted.
DORS obtains the highest PSNR and SSIM in
Tab.~\ref{tab:imagenet},
with a 0.44\,dB PSNR gain over the strongest baseline.
It ranks second in Q-Align and VQ-R1.

\noindent\textbf{Transfer from deblurring to super-resolution.}
The motion deblurring DORS model in Tab.~\ref{tab:deblur} is applied to
$\times4$ bicubic SR on ImageNet at $\sigma_y=0.025$
(Tab.~\ref{tab:sr_imag} (b)) without retraining. 
RAM uses its official SR checkpoints, and
 zero-shot methods retain their pretrained
denoisers or image priors. OFTSR and InvFusion retain their SR weights of Tab.~\ref{tab:sr_imag} (a), while only DORS uses the trained deblurring model.
DORS reaches 27.08\,dB, the highest value
among the compared methods by changing the operator during transfer.

\noindent\textbf{Transfer from single-coil to multi-coil MRI.}
We reuse the single-coil DORS model from
Sec.~\ref{sec:mri}, replacing $\mH=MF$ with
$\mH=MFS$ using ESPIRiT~\cite{uecker2014espirit} sensitivity maps $S$.
One global scale for the fusion coefficient $\rho_k$ is
calibrated on 128 training slices and fixed for testing.
PDNet retains its single-coil weights and applies its
measurement-domain layers separately to each coil.
RAM and MoDL use their officially released multi-coil weights.
uDPIR is unstable in this multi-coil setting, and the error grows across the unrolled iterations when directly changing operators, and rescaling the data-consistency weight does not help.
In Tab.~\ref{tab:mri} (b), RAM leaves residual error outside the coil support, lowering its whole-image scores. DORS achieves 36.75\,dB at $\times4$ and 33.47\,dB at
$\times8$, the highest PSNR at both accelerations, and the highest SSIM at $\times8$. Fig.~\ref{fig:mriiing} shows that with fixed refiner weights, DORS still stably reconstructs more content and detail, with sharper edges.

\dorsneedspace{4\baselineskip}
\subsection{Efficiency accounting}\label{app:efficiency}
Tab.~\ref{tab:efficiency} reports inference cost for $256\times256$ BSD100 deblurring (batch size $1$, strict FP32, same GPU) with the configurations of Tab.~\ref{tab:deblur}.
MACs include required backward computations and exclude FFT calls. We also report peak GPU memory (Mem.) and median inference time (Time) over images. 
% DU$^2$M's parameters are counted by its released code. 
With $16$ evaluations of its $3.7$M-parameter refiner, DORS has the fewest parameters and the lowest peak memory, and is more than $500\times$ faster than CLAMP. 
RAM and InvFusion use fewer MACs and run about twice as fast, with about $10\times$ and $46\times$ as many parameters.

% ==================================================================
\FloatBarrier
\section{Conclusion}\label{sec:conclusion}

We present DORS, a compact deterministic solver for linear inverse problems: analytic dual-observation fusion plus a shared 3.7M-parameter refiner trained through the complete chain. The propagated state admits an exact trajectory-center representation, its center fixed per step. We characterize how the measurement operator filters corrections toward this center, and controlled comparisons support its analytic use. Its placement matters beyond additional curvature. Experiments on four inverse problems show competitive performance. With fixed refiner weights, DORS transfers from deblurring to super-resolution and single-coil to multi-coil MRI. This supports explicit measurement updates and propagated states for compact solvers with learned refinement reusable across operators.
Limitations are provided in the Appendix.

\section*{LLM Usage Statement}

We used ChatGPT to assist with editing and optimizing the text of this paper, as well as debugging the code. All code and contents were reviewed and tested by the authors. We take responsibility for the final content of this work.

{
\small
\bibliographystyle{ieeenat_fullname}
\bibliography{references}
}
\end{document}